\documentclass{article}

\usepackage{arxiv}

\usepackage[utf8]{inputenc}
\usepackage[T1]{fontenc}
\usepackage{hyperref}
\usepackage{url}
\usepackage{booktabs}
\usepackage{amsfonts}
\usepackage{amsmath}
\usepackage{nicefrac}
\usepackage{microtype}
\usepackage{graphicx}
\usepackage{mdframed}
\usepackage{float}
\usepackage{caption}
\usepackage{amssymb}
\usepackage[round,authoryear]{natbib}
\setcitestyle{round}
\usepackage{algorithm}
\usepackage{algpseudocode}

\graphicspath{{./images/}}

\title{Near-Equivalent Q-learning Policies for Dynamic Treatment Regimes}

\author{
  Sophia Yazzourh \\
  Department of Epidemiology and Biostatistics\\
  McGill University\\
  Montréal, Québec, Canada \\
  \texttt{sophia.yazzourh@mcgill.ca} \\
  \And
  Erica E.M. Moodie \\
  Department of Epidemiology and Biostatistics\\
  McGill University\\
  Montréal, Québec, Canada \\
}

\begin{document}
\maketitle

\begin{abstract}
Precision medicine aims to tailor therapeutic decisions to the individual characteristics of each patient. This objective is commonly formalized through the framework of dynamic treatment regimes, which use statistical and machine learning methods to derive sequential decision rules adapted to evolving clinical information. In most existing formulations, these approaches produce a single optimal treatment recommendation at each stage, leading to a unique sequence of decision rules. However, in many clinical settings several treatment options may yield very similar expected outcomes, and restricting attention to a single optimal policy may conceal meaningful alternatives. In this work, we extend the Q-learning framework for retrospective data to incorporate a worst-value tolerance criterion controlled by a tuning hyperparameter $\varepsilon$, which specifies how far a policy is allowed to deviate from the optimal expected value. Rather than identifying a single optimal policy, the proposed approach constructs sets of $\varepsilon$-optimal policies whose expected value remains within a controlled neighborhood of the optimum. Conceptually, this formulation shifts the Q-learning problem from a vector-valued representation, in which a single value function determines the optimal decision, to a matrix-valued representation where multiple admissible value functions coexist during the backward recursion. This leads to families of near-equivalent treatment strategies instead of a single deterministic rule and explicitly characterizes regions of treatment indifference where several decisions achieve comparable expected outcomes. The proposed framework is illustrated through two settings. First, in a single-stage treatment decision problem, we examine how the admissibility criterion induces regions of treatment indifference around the decision boundary. Second, in a multi-stage treatment decision process, we apply the method within a sequential decision framework based on a simulated oncology model describing the joint evolution of tumor size and treatment toxicity.
\end{abstract}


\section{Introduction}\label{Introduction}

Precision medicine \citep{kosorok2019precision} places individual patient characteristics at the center of treatment decisions, with the goal of matching each patient with the treatment expected to confer the greatest clinical benefit. This question is particularly relevant in longitudinal settings with multiple decision points, as commonly encountered in the management of chronic diseases such as cancer, diabetes, or mental health disorders. Dynamic Treatment Regimes (DTRs) \citep{chakraborty2013statistical, kosorok2015adaptive} provide a formal framework for this paradigm by defining sequences of decision rules that adapt treatment recommendations to evolving patient characteristics. Over the last two decades, numerous methods have been proposed to estimate optimal DTRs, which can broadly be divided into two families. On the one hand, nonparametric methods such as inverse probability of treatment weighting \citep{robins2000marginal}, marginal structural models \citep{robins2000marginal, Orellana2010-zu}, and outcome weighted learning \citep{Zhao2012-vc} aim to directly learn optimal strategies without imposing strong parametric assumption. On the other hand, regression-based approaches including G-estimation \citep{robins1992estimation, robins1989analysis, robins1998correction}, dynamic weighted ordinary least squares \citep{wallace2015doubly}, and Q-learning \citep{clifton2020q} rely on statistical modeling to estimate how different treatment decisions would affect patient outcomes, thereby identifying strategies that maximize clinical benefits.

In this paper, we focus on Q-learning, a method rooted in reinforcement learning that frames treatment selection as a sequence of decision problems. At each stage, the method evaluates the expected benefit of each possible treatment, taking into account both immediate outcomes and their potential influence on future stages of care. By working backward from the final stage, these evaluations are combined to derive strategies that guide treatment choices in a way that maximizes overall patient benefit. Q-learning relies on regression models to estimate stage-specific value functions, which makes it particularly appealing in clinical applications. Parametric and semi-parametric modeling approaches are often familiar to practitioners, allow straightforward model selection and interpretation, and can be implemented using standard statistical tools while remaining flexible enough to capture complex treatment–covariate interactions. For these reasons, Q-learning has been widely adopted in precision medicine research and applied across diverse clinical contexts, as documented in review articles \citep{clifton2020q,yazzourh2025medical}.

Despite these advances, most existing methods are designed to identify a single optimal treatment policy. This "one-size-fits-all optimum" may be limiting in practice: clinical decisions often involve near-equivalent strategies, where multiple courses of action yield comparable expected outcomes. Restricting recommendations to a unique policy risks overlooking clinically acceptable alternatives, thereby reducing flexibility and clinician autonomy in the decision-making process.  

The integration of clinical expertise and domain knowledge into reinforcement learning algorithms has been increasingly recognized as an important direction for the application of dynamic treatment regimes in healthcare, both to improve the realism of treatment recommendations and to foster trust in data-driven decision tools \citep{holzinger2016interactive, maadi2021review, love2023should, holzinger2019causability, yazzourh2025medical}. Motivated by this perspective, we adapt Q-learning to identify sets of near-equivalent treatment policies by retaining treatment decisions whose expected value remains within an $\varepsilon$-threshold of the optimal policy. While multiple-policy approaches have been explored in online reinforcement learning settings \citep{fard2011non, tang2020clinician}, their reliance on direct interaction with the environment limits applicability in healthcare contexts where learning must typically rely on previously collected clinical data. A related multi-objective framework incorporating patient preferences has been proposed for offline settings \citep{lizotte2016multi}. In contrast, our approach is designed for retrospective clinical datasets under a single-objective paradigm, mapping out therapeutically equivalent treatment pathways while maintaining a primary focus on outcome optimization.

In this work, we modify the Q-learning framework to allow a controlled level of deviation from the estimated optimal value. This makes it possible to identify several treatment actions that achieve similar performance. Our method introduces an $\varepsilon$-selection step during the backward recursion, which allows the retention of all treatment decisions whose estimated value remains within an $\varepsilon$-neighborhood of the optimum. Conceptually, this modification transforms the classical vector-based formulation of Q-learning into a matrix-based representation in which multiple admissible value functions are propagated through the recursion, ultimately producing sets of near-equivalent policies rather than a single deterministic rule.

The remainder of the paper is organized as follows. Section~2 briefly reviews the classical Q-learning framework. Section~3 introduces the proposed near-equivalent Q-learning approach and describes the $\varepsilon$-selection procedure. Section~4 illustrates the method in a single-stage setting by examining individualized treatment rules and the resulting decision boundaries. Section~5 considers a multi-stage dynamic treatment regime in a simulated oncology setting. Section~6 concludes with a discussion of the methodological implications and potential extensions.

\section{Q-learning}

Longitudinal clinical data consist of measurements collected sequentially throughout the course of treatment.
At baseline, they capture patient covariates such as demographic and clinical characteristics 
(e.g., age, sex, comorbidities, or genetic markers). As treatment progresses, some variables are reassessed at follow-up visits and may evolve over time, reflecting both the patient's response to therapy and the natural progression of the disease. At each stage, outcomes are recorded to quantify the benefit of the 
treatment received up to that point. Together, these data describe a dynamic trajectory in which 
the patient’s state evolves in response to past decisions and, in turn, shapes future treatment choices.

In the context of precision medicine, a Dynamic Treatment Regime provides a formal framework 
for individualizing care over time. We consider a setting with a finite number of decision stages, 
indexed by $t=0,\dots,T$, where treatment choices are made at discrete points during the course of care. 
A DTR is defined as a sequence of decision rules, each specifying how treatment should be adapted at a 
given stage according to the patient’s history up to that point. Formally, let $\mathcal{H}_t$ denote the 
history space available prior to stage $t$, and let $H_t \in \mathcal{H}_t$ denote the corresponding 
random history. Similarly, let $\mathcal{A}_t$ denote the treatment space at stage $t$, and 
$A_t \in \mathcal{A}_t$ the assigned treatment. The outcome subsequently observed is denoted by $Y_t$. The collection $\pi = \{\pi_0, \dots, \pi_T\}$ defines a DTR, where each decision rule is a mapping 
$\pi_t : \mathcal{H}_t \to \mathcal{A}_t$. For a realized history $h_t \in \mathcal{H}_t$, the rule 
assigns the treatment $a_t = \pi_t(h_t)$.

Several methodological frameworks have been developed to estimate optimal DTRs, and these can broadly be divided into two families. The first family focuses on directly optimizing the value of a regime, defined as the expected cumulative outcome under a given sequence of decisions; the idea is to evaluate how well a candidate strategy would perform in the population without explicitly modeling the outcome process at each stage. Within this framework, inverse probability weighting and marginal structural models \citep{robins2000marginal, Orellana2010-zu} use weighting schemes to correct for time-dependent confounding, while outcome weighted learning \citep{Zhao2012-vc} formulates the problem as a weighted classification task. Extensions of this approach include backward outcome weighted learning \citep{zhao2015new}, which adapts the method to multi-stage settings through backward recursion, and residual weighted learning \citep{zhou2017residual}, which replaces raw outcomes by residuals to improve stability and reduce variance. Despite their differences, these approaches share the common feature of targeting the regime value directly rather than modeling outcomes sequentially.

The second family is based on regression formulations, where sequential models are fit to capture the relationship between patient histories, treatment assignments, and outcomes. 
G-estimation \citep{robins1992estimation, robins1989analysis, robins1998correction}, dynamic weighted least squares, \citep{wallace2015doubly}, and Q-learning \citep{clifton2020q} all focus on recovering optimal strategies through sequential regression modeling of outcomes parameterized as structural nested mean models. 
In each approach, the analytic problem is conceived as a sequence of regression tasks solved backward in time, where each step predicts the cumulative outcome under alternative treatment choices. These methods derive optimal decision rules from recursively specified outcome models.

Within this regression-based family, Q-learning occupies a central place and will be the focus of what follows. Beyond its popularity in applied work, it is conceptually appealing because it makes explicit the connection between sequential decision problems in medicine and reinforcement learning \citep{clifton2020q, yazzourh2025medical}. Originally developed in the context of interactive online data, Q-learning was designed to learn strategies by both exploiting accumulated knowledge and exploring new actions to improve performance \citep{Watkins1989-er, watkins1992q}. At the heart of this framework is the notion of a reward, denoted by $R$ or $Y$, which provides a quantitative measure of the consequences of each decision; in medical applications, this notion coincides with the observed outcome. In Q-learning, the central objects are the $Q$-functions, which quantify the expected cumulative reward (i.e outcome) 
associated with taking a given action in a given state under a regime $\pi$. Intuitively, they measure how beneficial it is for a patient $i$ with history $h_{i,t}$
to choose treatment $a_{i,t}$ at time $t$, while following strategy $\pi$ thereafter. For patient $i \in \{1,\dots,N\}$, the stage-$t$ $Q$-function is defined as
\[
Q_{i,t}^\pi(h_{i,t},a_{i,t}) = \mathbb{E}^\pi\!\left[\sum_{t=0}^T Y_{i,t}(h_{i,t},a_{i,t}) \,\middle|\, H_{i,t}=h_{i,t}, A_{i,t}=a_{i,t} \right].
\]

In the context of DTR, the data used for estimation consist of 
observations that have already been collected in clinical settings, either through formal 
clinical trial protocols or through medical decisions made in practice. In reinforcement 
learning, such datasets are considered \emph{offline}, and the associated $Q$-functions are 
estimated retrospectively from pre-existing observations.As described in Algorithm~\ref{alg:BQ}, the procedure begins by estimating the $Q$-function 
at the final stage $T$ and then proceeds backward in time to earlier stages. This approach 
is commonly referred to as offline or backward Q-learning.

In practice, the true $Q$-functions are unknown and must be estimated from the data. Let 
$\hat{Q}_t$ denote the estimator of $Q_t$, obtained by fitting a regression model that predicts future 
outcomes from patient histories and treatments. Such estimators may rely on parametric models 
(e.g., linear or generalized linear models) or on more flexible machine learning methods, including 
support vector machines or deep neural networks.

At the final stage $T$, the $Q$-function is estimated by regressing the final outcome on the covariates and treatment. For earlier stages $t < T$, the regression target is no longer the observed outcome 
alone but a modified outcome that incorporates the value of the best possible future 
decision. This captures how treatment choices at time $t$ influence downstream rewards 
and links the decision process across time. At each stage, an optimal treatment is obtained by selecting the action associated with the 
largest estimated $Q$-value, where a $Q$-value represents the expected cumulative outcome for a given patient history $h_t$ and treatment choice $a$ at stage $t$, assuming optimal decisions are followed thereafter. 
In practice, these $Q$-values are computed by evaluating the estimated $Q$ function at the specific history and treatment pair $(h_t,a)$.

This construction is closely related to the potential outcomes framework where $Y$ denotes the observed outcome under the treatment actually received, whereas $Y^*(a)$ represents the counterfactual outcome that would have been observed under an alternative treatment $a$. In the recursive estimation procedure, the observed outcome at stage $t$ is replaced by a pseudo-outcome constructed from the estimated model at stage $t+1$, representing the optimal expected future values. Using this learned model, we evaluate all feasible future actions (i.e., all counterfactual treatment choices) and retain the one producing the highest value.

For the entire sample of $N$ patients, we write the modified outcome in vector form as
\[
\tilde{\mathbf{y}}_{t}
=
\begin{bmatrix}
\tilde{Y}_{1,t} \\
\tilde{Y}_{2,t} \\
\vdots \\
\tilde{Y}_{N,t}
\end{bmatrix}
=
\begin{bmatrix}
Y_{1,t} + \max_{a_{t+1}} \hat{Q}_{t+1}(H_{1,t+1}, a_{t+1}) \\
Y_{2,t} + \max_{a_{t+1}} \hat{Q}_{t+1}(H_{2,t+1}, a_{t+1}) \\
\vdots \\
Y_{N,t} + \max_{a_{t+1}} \hat{Q}_{t+1}(H_{N,t+1}, a_{t+1})
\end{bmatrix}.
\]

This representation highlights that for each patient, the unobserved continuation of their trajectory 
is replaced by the value they would have obtained under the optimal decision at the next stage. By 
propagating these optimal counterfactuals backward through time, Q-learning incorporates long-term and 
delayed treatment effects into the estimation.

Once all the $Q$-functions have been estimated across stages, the optimal strategy is obtained by 
selecting, at each stage $t$, the action that maximizes the corresponding estimated $Q$-function:
\[
\hat{\pi}^\star_t(h_t) = \arg\max_{a_t \in \mathcal{A}_t} \, \hat{Q}_t(h_t,a_t).
\]

\medskip

\noindent\textbf{Algorithm 1.Backward Q-learning}
\label{alg:BQ}

\vspace{0.3em}
\hrule
\vspace{0.5em}

\begin{mdframed}[
  linewidth=0.8pt,
  topline=false,
  skipabove=0pt,
  skipbelow=0pt,
  innerleftmargin=10pt,
  innerrightmargin=10pt
]
\begin{algorithmic}[1]

\Statex \textbf{Input}
\State Offline training data consisting of admissible histories $H_t$, treatments $A_t$, and outcomes $Y_t$ for $t=0,\dots,T$, and a fitting algorithm (parametric or non-parametric).

\Statex \textbf{Final stage $T$}
\State Estimate $Q_T$ from $(Y_T,H_T,A_T)$ such that
\[
\hat{Q}_T(h_T,a_T) \triangleq \mathbb{\hat{E}}\!\left(Y_T \mid H_T=h_T, A_T=a_T\right).
\]

\For{$t = T{-}1,\dots,0$}
  \State Compute the pseudo-outcome
  \[
  \tilde{Y}_t = Y_t + \max_{a_{t{+}1}} \hat{Q}_{t{+}1}(H_{t{+}1}, a_{t{+}1}).
  \]
  \State Estimate
  \[
  \hat{Q}_t(h_t,a_t) \triangleq \mathbb{\hat{E}}\!\left(\tilde{Y}_t \mid H_t=h_t, A_t=a_t\right).
  \]
\EndFor

\Statex \textbf{Output}
\State The estimated optimal policies $\{\hat{\pi}^\star_0,\dots,\hat{\pi}^\star_T\}$ with
\[
\hat{\pi}^\star_t(h_t) = \arg\max_{a_t\in\mathcal{A}_t} \hat{Q}_t(h_t,a_t).
\]

\end{algorithmic}
\end{mdframed}

\section{Near-equivalent Q-learning}

In its standard formulation, Q-learning recommends a single action at each decision point, namely the one that maximizes the estimated expected outcome at time $t$. While this yields a clear decision rule, it may conceal alternative actions, or even entire treatment sequences, that achieve comparable expected outcomes.

This limitation is particularly salient in clinical applications. Treatment policies are typically learned from observational or trial data generated under pre-existing medical strategies, often shaped by physician judgment or fixed study protocols. Consequently, the estimated policy reflects not only the outcome process but also the historical allocation mechanism embedded in the data. The resulting recommendation therefore inherits the exploration–exploitation structure of the data-generating process.

Restricting the analysis to a unique optimal strategy may thus reduce flexibility and obscure near-equivalent treatment pathways that would be clinically acceptable. When estimated value differences are small, the distinction between optimal and suboptimal actions may be driven more by statistical variability than by meaningful clinical contrast. Identifying strategies with comparable expected performance may therefore provide a more informative and practically relevant characterization of the decision landscape.

One possible way to address the limitations of single-policy Q-learning is to construct a framework that recommends \emph{near-equivalent actions}, thereby allowing the integration of additional considerations such as treatment invasiveness, local availability, or potential side effects. Rather than enforcing strict optimality, such an approach seeks to identify actions whose expected performance remains within a controlled margin of the optimal value.

From a decision-process perspective, this idea is closely related to the concept of \emph{near-optimal policies} introduced by \citeauthor{fard2011non} \citeyear{fard2011non}. In that work, admissible actions are defined through a \emph{worst-case value} criterion, corresponding to the smallest $Q$-value retained within the acceptable set. This quantity determines both the permissible deviation from the optimal decision and the minimal performance level tolerated among admissible actions. The resulting framework yields a set of policies whose performance is guaranteed to remain within a predefined threshold.

However, the approach of \citeauthor{fard2011non} \citeyear{fard2011non} was developed in a fully model-based setting, assuming known transition probabilities, and within an interactive (on-policy) environment where the policy is optimized through direct interaction with the system. These assumptions differ substantially from real-world Dynamic Treatment Regime applications, which rely on observational or trial data and do not permit interactive policy refinement.

The method was subsequently extended by \citeauthor{tang2020clinician} \citeyear{tang2020clinician} to a model-free context and an off-policy setting, where the environment is first explored and then exploited in a separate optimization phase. Although this extension relaxes the requirement of known transition dynamics, it remains situated within an online learning paradigm and was evaluated in a simulated intensive-care framework based on MIMIC data. As such, it does not align with the constraints of offline DTR estimation.

Finally, the notion of multiple admissible policies has also been explored in a multi-objective reinforcement learning framework by \citeauthor{lizotte2016multi} \citeyear{lizotte2016multi}. In that setting, the reward is vector-valued and reflects several competing clinical objectives. The authors introduce a dominance concept of Pareto dominance, incorporating patient preferences. Applied to the CATIE study, this approach enables the identification of treatment strategies that balance multiple outcomes while remaining clinically interpretable.

While conceptually related, existing approaches typically operate in model-based, interactive, or multi-objective settings. In contrast, our objective is to introduce controlled near-optimal admissibility within a single-objective, offline, regression-based Dynamic Treatment Regime framework. To this end, we consider a finite-horizon DTR setting in which $Q$-functions are estimated via classical backward Q-learning. Let $T$ denote the terminal stage. As described previously, standard backward Q-learning proceeds recursively from stage $T$ to stage $0$, assuming that the optimal action is taken at subsequent stages when constructing pseudo-outcomes. Our goal is to embed near-optimal admissibility within this regression-based recursion while preserving its structural integrity.

The admissibility criterion introduced by \citeauthor{fard2011non} \citeyear{fard2011non} is originally formulated in terms of value functions, which are assumed to be positive. In their framework, a subset of actions $S \subseteq \mathcal{A}_t$ is considered $\varepsilon$-admissible if
\[
\min_{a \in S} V(h_t,a)
\ge
(1-\varepsilon)\max_{a \in \mathcal{A}_t} V(h_t,a).
\]
This definition relies on a worst-case perspective: the minimum value within the admissible set must remain within a $(1-\varepsilon)$ margin of the optimal (i.e., maximal) value, ensuring that even the least favorable admissible action remains close to the optimum.

Applying this condition directly within a Q-learning framework would require imposing the same criterion on the estimated Q-functions. However, unlike the value functions considered in the original formulation, Q-functions are not necessarily positive. Consequently, the multiplicative formulation above may lead to undesirable behavior when $\max_{a \in \mathcal{A}_t}\hat{Q}_t(h_t,a)$ is negative. 

To address this issue, we adopt the following admissibility condition. For a given history $h_t$, an action $a \in \mathcal{A}_t$ is considered admissible if
\[
\hat{Q}_t(h_t,a)
\ge
\max_{a \in \mathcal{A}_t}\hat{Q}_t(h_t,a)
-
\varepsilon
\left|
\max_{a \in \mathcal{A}_t}\hat{Q}_t(h_t,a)
\right|.
\]
This condition ensures that the estimated value of any retained action lies within an $\varepsilon$-neighborhood of the optimal estimated value. Equivalently, the maximal tolerated loss relative to the optimal action is bounded by $\varepsilon\left|\max_{a \in \mathcal{A}_t}\hat{Q}_t(h_t,a)\right|$. The hyperparameter $\varepsilon$ therefore controls the tolerated deviation from optimality. When $\varepsilon = 0$, the procedure reduces to classical Q-learning, whereas increasing values of $\varepsilon$ progressively enlarge the set of near-optimal admissible actions.

The role of $\varepsilon$ can be further clarified by examining its boundary cases. When $\varepsilon = 0$, the admissibility condition reduces to
\[
\hat{Q}_t(h_t,a)
\ge
\max_{a \in \mathcal{A}_t}\hat{Q}_t(h_t,a),
\]
which implies that only actions whose estimated value is strictly equal to the maximal estimated value are retained.

At the other extreme, when $\varepsilon = 1$, the condition becomes
\[
\hat{Q}_t(h_t,a)
\ge
\max_{a \in \mathcal{A}_t}\hat{Q}_t(h_t,a)
-
\left|
\max_{a \in \mathcal{A}_t}\hat{Q}_t(h_t,a)
\right|.
\]

Two situations may arise. If the estimated Q-functions are strictly positive, the condition reduces to $\hat{Q}_t(h_t,a) \ge 0$. One might therefore expect that all strategies would be retained when rewards are strictly positive. However, in practice, Q-functions are estimated through regression models that do not constrain predictions to remain within the range of the observed outcomes. Many regression procedures allow extrapolation beyond the support of the target variable. For example, in the second implementation presented in Section~5, we use a support vector regression model with a radial basis function (Gaussian) kernel, which does not enforce positivity of predictions even when all observed rewards are positive. More generally, unless explicit constraints are imposed, estimated Q-values may take negative values. Consequently, $\varepsilon = 1$ does not guarantee that all strategies are retained, and the admissibility criterion becomes ill-posed at that boundary. If the maximal estimated Q-value is negative, the condition also becomes difficult to interpret, since the admissibility threshold may no longer correspond to a clear notion of near-optimality. For these reasons, we restrict the hyperparameter to $\varepsilon \in [0,1[$ to ensure a well-defined and controlled relaxation of optimality.

The hyperparameter $\varepsilon$ therefore controls the maximal loss tolerated with respect to the optimal treatment. Importantly, this loss is defined relative to the magnitude of the maximal estimated value. In other words, the admissibility threshold is not defined by a fixed distance from the optimum, but by a proportion of the size of the best estimated effect. When $\max_{a \in \mathcal{A}_t}\hat{Q}_t(h_t,a)$ is large, the admissible band becomes wider, reflecting the fact that several actions may yield similar large expected benefits. Conversely, when the maximal estimated value is small, the admissible band becomes narrower, meaning that only actions very close to the optimum are retained.

Under this admissibility criterion, the decision rule at stage $t$ becomes set-valued rather than single-valued, mapping each history $h_t$ to a subset of $\mathcal{A}_t$ containing all actions whose estimated value lies within the prescribed tolerance of the optimum. The sequential composition of these admissible sets across stages therefore defines a collection of near-equivalent treatment trajectories instead of a unique strategy. A necessary condition for implementation is that the action space $\mathcal{A}_t$ be finite and discrete, allowing explicit comparison across actions; in DTR applications, this assumption is natural, as treatment regimens are clinically predefined and finite.

A direct application of $\varepsilon$-selection at every stage would disrupt the recursive structure of backward Q-learning. Standard recursion relies on the assumption that a unique optimal action is taken at future stages when constructing pseudo-outcomes. Allowing set-valued actions at all stages would therefore break this dependency structure and lead to a combinatorial expansion of candidate trajectories. Thus, to preserve the classical backward scheme, we introduce $\varepsilon$-selection only at the penultimate stage. Applying the selection at stage $T-1$ allows identification of near-equivalent treatment trajectories while preserving a unique optimal action at the terminal stage. If selection were introduced at the terminal stage alone, potentially suboptimal deviations could propagate backward and accumulate error. By contrast, selecting at the second-to-last stage captures multiple clinically comparable strategies while maintaining consistency with the estimated optimal outcome at $T$.

At stage $T$, a single $\hat{Q}_T$ is estimated. When $\varepsilon$-selection is applied at stage $T-1$, for each patient $i$, it yields a set of $n_i$ admissible $Q$-functions, denoted
$$
Q^1_{i,T}, \dots, Q^{n_i}_{i,T}.
$$

\noindent Let
$$
m=\max_{i \in \{0,\dots,N\}}(n_i), \qquad m \leq |\mathcal{A}_T|.
$$

\noindent Since regression at stage $T-2$ relies on pseudo-outcomes derived from stage $T-1$, these quantities must share a common dimension across individuals in order to fit a single regression model and maintain a coherent recursive structure. Let $m$ denote the maximum number of admissible actions observed at stage $T-1$. For each individual, we therefore construct a pseudo-outcome vector of length $m$. When $n_i < m$, the vector is completed by padding with repeated evaluations of the first optimal $Q$-function until dimension $m$ is reached. This construction yields a square and uniformly dimensioned pseudo-outcome matrix at stage $T-2$, simplifying the backward recursion at earlier stages and avoiding stage-specific model specifications. Importantly, the padding operation neither alters the relative ordering of admissible actions nor changes the long-term optimization objective; it preserves the value structure induced by the estimated $Q$-functions while ensuring the dimensional alignment required for stable and consistent recursive estimation. 

Consequently, for stages $t < T-1$, the pseudo-outcome no longer takes the form of a vector as in Section~2, but becomes a matrix $\tilde{\mathbf{Y}}_{t}$ of dimension $(N \times m)$, where $m$ denotes the maximum number of per-individual $Q$-functions retained under the $\varepsilon$-selection criterion at stage $T$. Specifically,
$$
\tilde{\mathbf{Y}}_{t}
=
\begin{pmatrix}
\tilde{Y}^1_{1,t} & \dots & \tilde{Y}^m_{1,t} \\
\vdots & \ddots & \vdots \\
\tilde{Y}^1_{N,t} & \dots & \tilde{Y}^m_{N,t}
\end{pmatrix},
$$
where, for $j \in \{1,\dots,m\}$,
$$
\tilde{Y}^j_{i,t}(h_{i,t})
=
Y_{i,t}(h_{i,t})
+
\max_a \hat{Q}^j_{i,t+1}(h_{i,t+1}, a).
$$
For each $j$, a separate regression model is fitted using the pseudo-outcome vector $\boldsymbol{\tilde{y}}^j_t$ (the $j$-th column of $\tilde{\mathbf{Y}}_{t}$), yielding the corresponding estimator $\hat{Q}^j_t$. The backward recursion therefore proceeds as in classical Q-learning, except that it is carried out in parallel across the $m$ retained models. The complete procedure is summarized in Algorithm~\ref{alg:NearQ} and produces a collection of $\varepsilon$-equivalent treatment strategies within a standard regression-based DTR framework. Rather than discarding optimality, the method relaxes it in a controlled manner governed by the hyperparameter $\varepsilon$, thereby preserving the long-term optimization objective while allowing multiple near-optimal treatment trajectories.

Clinically, this approach enables identification of patients located near decision boundaries, for whom multiple actions yield similar expected outcomes. For such patients, external considerations such as toxicity, availability, cost, or patient preference may legitimately guide the final choice without materially compromising expected performance. Methodologically, the approach remains fully grounded in backward Q-learning, requiring only a modified pseudo-outcome construction at stage $T-1$.

In the following section, we evaluate the proposed method in two complementary settings. 
We first consider a simulated Individualized Treatment Regime scenario (i.e., a single-stage decision process setting) with binary treatment, which allows for a clear investigation of the role of the hyperparameter $\varepsilon$ and highlights the ability of the method to detect patients located near the decision boundary. 
We then illustrate the approach in a multi-stage DTR setting with six treatment stages in a simulated oncology framework to assess its practical performance and potential clinical relevance. 
All code and simulation results are publicly available on a dedicated GitHub repository\footnote{\url{https://github.com/sophiayazzourh/Multiple_Policies_Qlearning}}.

\noindent\textbf{Algorithm 2. Near-Equivalent Q-learning}
\label{alg:NEQ}

\vspace{0.3em}
\hrule
\vspace{0.5em}

\begin{mdframed}[
  linewidth=0.8pt,
  topline=false,
  skipabove=0pt,
  skipbelow=0pt,
  innerleftmargin=10pt,
  innerrightmargin=10pt
]
\begin{algorithmic}[1]

\Statex \textbf{Input}: Offline training data consisting of $i \in \{1,\dots,N\}$ patients, each with admissible histories $h_{i,t}$ and associated rewards $Y_{i,t}$ for $t \in \{0,\dots,T\}$. The algorithm also requires a fitting method (parametric or non-parametric) and a hyperparameter $\varepsilon \in [0;1[$ that specifies the allowed deviation from the optimal policy.

\Statex \textbf{Final stage $T$:}

\underline{Estimation}: Estimate $Q_T$ based on $Y_T \sim H_T, A_T$, such that
$$
\hat{Q}_T(h_T, a_T) \triangleq \mathbb{\hat{E}}(Y_T \mid H_T = h_T, A_T = a_T).
$$
At this stage, we obtain a single estimated $Q$-function, denoted $\hat{Q}_T$.

\Statex \textbf{Penultimate stage $T-1$}:
\begin{enumerate}
    \item \underline{$\varepsilon$-Selection}: for each patient $i$, identify the subset of estimated $Q_T$-functions whose value satisfies the $\varepsilon$-admissibility condition,
    $$
    \hat{Q}_{i,T}(h_{i,T}, a)
    \ge
    \max_{a \in \mathcal{A}_T} \hat{Q}_{i,T}(h_{i,T}, a)
    -
    \varepsilon
    \left|
    \max_{a \in \mathcal{A}_T} \hat{Q}_{i,T}(h_{i,T}, a)
    \right|.
    $$
    Let $n_i$ denote the number of estimated $Q$-functions satisfying this criterion for patient $i$.  
    We then construct the vector
    $$
    \left[
    \hat{Q}^{1}_{i,T}, \dots, \hat{Q}^{n_i}_{i,T}
    \right],
    $$
    containing the admissible estimated $Q$-values for patient $i$.
   \item \underline{Padding}: Let $m = \max_{i \in \{1,\dots,N\}}(n_i)$ denote the maximum number of per-individual $Q$-functions satisfying the $\varepsilon$-selection criterion at stage $T$, where $m \leq |\mathcal{A}_T|$ and $|\mathcal{A}_T|$ denotes the cardinality of the action space. We construct an $(N \times m)$ matrix in which the $i$-th row consists of the $n_i$ selected elements $[\hat{Q}^{1}_{i,T}, \dots, \hat{Q}^{n_i}_{i,T}]$, followed, when $n_i < m$, by $(m - n_i)$ copies of $\hat{Q}^{1}_{i,T}$ to ensure common dimensionality across patients.
    This results, for each patient $i$, in a vector of dimension $(1 \times m)$ given by
    $$
    \left[
    \hat{Q}^{1}_{i,T}, \dots, \hat{Q}^{m}_{i,T}
    \right].
    $$
    
    \item \underline{Pseudo-outcome matrix}: construct the pseudo-outcome matrix $\tilde{\mathbf{Y}}_{T-1}$ of dimension $(N \times m)$:
$$
\tilde{\mathbf{Y}}_{T-1}
=
\begin{pmatrix}
\tilde{Y}^1_{1,T-1} & \dots & \tilde{Y}^m_{1,T-1} \\
\vdots & \ddots & \vdots \\
\tilde{Y}^1_{N,T-1} & \dots & \tilde{Y}^m_{N,T-1}
\end{pmatrix},
$$
where, for $j \in \{1,\dots,m\}$,
$$
\tilde{Y}^j_{i,T-1}(h_{i,T-1})
=
Y_{i,T-1}(h_{i,T-1})
+
\max_a \hat{Q}^j_{i,T}(h_{i,T}, a).
$$

\item \underline{Estimation}: Estimate the functions $Q^j_{T-1}$ by fitting separate regressions to each column of $\tilde{\mathbf{Y}}_{T-1}$. 
Let $\tilde{\mathbf{y}}^j_{T-1}$ denote the pseudo-outcome vector associated with model $j$, i.e., the $j$-th column of $\tilde{\mathbf{Y}}_{T-1}$. For $j = 1,\dots,m$, we define
$$
\hat{Q}^j_{T-1}(h_{T-1}, a_{T-1})
\triangleq
\mathbb{\hat{E}}\big(
\tilde{\mathbf{y}}^j_{T-1}
\mid
H_{T-1} = h_{T-1}, 
A_{T-1} = a_{T-1}
\big).
$$
At stage $T-1$, this yields a collection of $m$ optimal and near-optimal $Q$-functions:
$$
\left\{
\hat{Q}^{1}_{T-1}, \dots, \hat{Q}^{m}_{T-1}
\right\}.
$$
\end{enumerate}

\Statex \textbf{Stages $t < T-1$}:
\begin{enumerate}
    \item \underline{Pseudo-outcome matrix}: construct the pseudo-outcome matrix $\tilde{\mathbf{Y}}_{t}$ of dimension $(N \times m)$:
    $$
    \tilde{\mathbf{Y}}_{t}
    =
    \begin{pmatrix}
    \tilde{Y}^1_{1,t} & \dots & \tilde{Y}^m_{1,t} \\
    \vdots & \ddots & \vdots \\
    \tilde{Y}^1_{N,t} & \dots & \tilde{Y}^m_{N,t}
    \end{pmatrix},
    $$
    where, for $j = 1,\dots,m$,
    $$
    \tilde{Y}^j_{i,t}(h_{i,t})
    =
    Y_{i,t}(h_{i,t})
    +
    \max_a \hat{Q}^j_{i,t+1}(h_{i,t+1}, a).
    $$ 

    \item \underline{Estimation}: For $j = 1,\dots,m$, estimate the functions $Q^j_{t}$ by fitting separate regressions to each column of $\tilde{\mathbf{Y}}_{t}$. 
    Let $\tilde{\mathbf{y}}^j_{t}$ denote the pseudo-outcome vector associated with model $j$, i.e., the $j$-th column of $\tilde{\mathbf{Y}}_{t}$. We define
    $$
    \hat{Q}^j_{t}(h_{t}, a_{t})
    \triangleq
    \mathbb{\hat{E}}\big(
    \tilde{\mathbf{y}}^j_{t}
    \mid
    H_{t} = h_{t}, 
    A_{t} = a_{t}
    \big).
    $$
    
    At each stage $t$, this yields a collection of $m$ optimal and near-optimal $Q$-functions:
    $$
    \left\{
    \hat{Q}^{1}_{t}, \dots, \hat{Q}^{m}_{t}
    \right\}.
    $$
\end{enumerate}

\Statex \textbf{Construction of $\varepsilon$-equivalent estimated policy sets $\hat{\Pi}^T$:}
At each stage $t$, we construct a set of near-equivalent actions. 
The sequential combination of these actions defines a set of near-equivalent strategies such that, for $j \in \{1,\dots,m\}$,
$$
\hat{\Pi}^{j}(h)
=
\left[
\hat{\pi}^{j}_0(h_0), \dots, \hat{\pi}^{j}_T(h_T)
\right],
$$
where
$$
\hat{\pi}^{j}_t(h_t)
=
\arg\max_{a_t \in \mathcal{A}_t}
\hat{Q}^{j}_t(h_t, a_t).
$$

\end{algorithmic}
\end{mdframed}

\section{Individualized Treatment Regimes: decision boundaries and the role of $\varepsilon$}
\label{sec:itr_boundary}
We consider a single-stage individualized treatment regime and simulate data according to \citep{song2015penalized}. For each simulated patient, covariates $X_0,\dots,X_9$ are generated independently and identically distributed as $\mathcal{U}([-1,1])$, and $A \in \{-1,1\}$ with $\mathbb{P}(A=1)=1/2$. The outcome is generated as $Y \sim \mathcal{N}\!\big(1 + 2X_0 + X_1 + 0.5X_2 + T_0(A,X),\,1\big)$, where $T_0(A,X) = (X_0 + X_1)A$. The true individualized treatment effect is $\tau(X)=X_0+X_1$, the optimal rule is $\pi^\star(X)=\mathrm{sign}(X_0+X_1)$, and the true decision boundary is the linear surface $X_0+X_1=0$.


We implement a one-stage Q-learning procedure using a linear working model for the $Q$-function.
Specifically, we estimate
\[
\widehat{Q}(X,A)
=
\widehat{\beta}_0
+ \widehat{\beta}_X^\top X
+ \widehat{\beta}_A A
+ \widehat{\beta}_{AX}^\top (A X),
\]
where treatment--covariate interaction terms allow for heterogeneous treatment effects. The model is fitted on the training sample using a linear regression with regressors $[1, X_0,\dots,X_9, A, A X_0,\dots, A X_9]$, i.e., including all covariates and all treatment--covariate interaction terms.
For each subject in the test set, predicted values are computed under both treatment options $A=-1$ and $A=+1$, yielding the estimated value pair $\bigl(\widehat{Q}(X,-1), \widehat{Q}(X,+1)\bigr).$


In the one-stage binary-treatment setting, the role of $\varepsilon$ is naturally understood through the blip function, defined as the difference between the $Q$-values evaluated under each of the two treatment options, $\Delta(X) = Q(X,1) - Q(X,-1)$. The blip fully determines the individualized treatment rule: its sign specifies the optimal action, while its magnitude reflects the strength of the treatment preference. From a reinforcement learning perspective, $\Delta(X)$ corresponds to a relative advantage function, quantifying how much better one action performs compared to the alternative at the same covariate profile. 

In this setting, $\varepsilon$-selection is constructed directly from the estimated blip $\widehat{\Delta}(X) = \widehat{Q}(X,1) - \widehat{Q}(X,-1)$ by retaining both treatments whenever $|\widehat{\Delta}(X)| \le \varepsilon$, equivalently including all actions whose predicted value lies within $\varepsilon$ of the maximal predicted $Q$-value. Geometrically, this replaces the sharp decision boundary $\widehat{\Delta}(X)=0$ with a band defined by $|\widehat{\Delta}(X)| \le \varepsilon$. As $\varepsilon$ increases, this band widens, identifying regions of the covariate space where the difference in expected outcomes between treatments is small and where no clearly dominant choice emerges.

\begin{figure}[ht]
\centering
\includegraphics[width=0.9\textwidth]{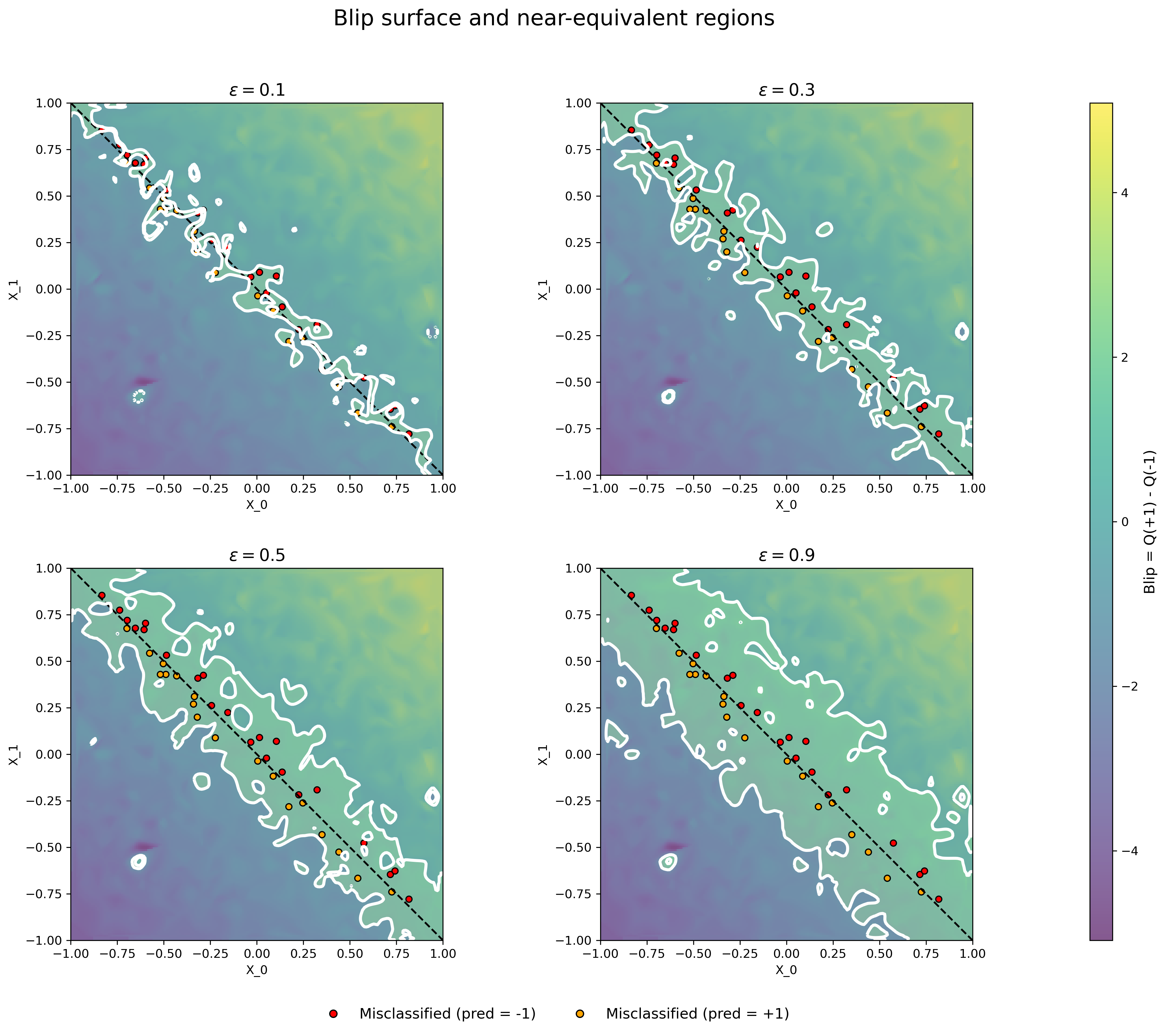}
\caption{Estimated blip surface $\widehat{\Delta}(X) = \widehat{Q}(X,1) - \widehat{Q}(X,-1)$ projected onto the $(X_0,X_1)$ plane for increasing values of $\varepsilon$. The dashed line represents the true decision boundary $X_0 + X_1 = 0$. White contours indicate the $\varepsilon$-selection region defined by $|\widehat{\Delta}(X)| \le \varepsilon$.}
\label{fig:itr_eps}
\end{figure}


Figure~\ref{fig:itr_eps} displays the estimated blip surface projected onto the $(X_0, X_1)$ plane. Positive regions (yellow) indicate preference for treatment $+1$, while negative regions (purple) favor treatment $-1$. The smooth transition of the surface reflects the linear treatment--covariate interaction used in the data-generating mechanism.

Misclassified points are primarily located near the true decision boundary, where the treatment effect is close to zero and small estimation fluctuations may alter the sign of the blip. The white band corresponds to the $\varepsilon$-selection region defined by $|\widehat{\Delta}(h)| \le \varepsilon$. This band encompasses most misclassified observations and identifies regions where the expected difference between treatments is small. Rather than imposing a sharp boundary, the $\varepsilon$-selection framework explicitly acknowledges that, in these areas, no clearly dominant treatment emerges. This perspective is related in spirit to earlier approaches that address instability near the decision boundary through hard- or soft-thresholding \citep{moodie2010estimating, chakraborty2010inference}, although the present framework differs in that it yields a set of admissible treatments rather than modifying the decision rule itself.

In contrast, misclassifications are rare in regions where the blip has large magnitude, indicating that the model reliably identifies the optimal treatment when the signal is strong. Overall, the concentration of apparent errors near the $\varepsilon$-band suggests that they reflect intrinsic ambiguity in treatment preference rather than model inadequacy.

\section{Cancer Simulation Study: A Multistage Multiple-Treatment Decision Problem}

The goal of this section is to illustrate the behavior of the proposed approach in a setting with multiple treatments and multiple stages. As a point of reference, we consider standard Q-learning, since our method is explicitly designed as an extension of it, modifying only the decision rule while preserving the underlying estimation procedure. Existing approaches for constructing set--valued or near--optimal treatment rules are typically developed in model-based, interactive, or multi--objective settings \citep{fard2011non, lizotte2016multi, tang2020clinician}, and do not directly accommodate the offline, regression-based Dynamic Treatment Regime framework considered here. Consequently, there are currently no direct comparable methods that address the same data structure.

We illustrate the proposed method using a simulated dataset based on the model proposed by \citeauthor{zhao2009reinforcement}~(\citeyear{zhao2009reinforcement}), a benchmark frequently used in reinforcement learning studies \citep{furnkranz2012preference, akrour2012april, goldberg2012q, humphrey2017using, yazzourh2024reinforcement}. The model represents a simplified form of chemotherapy management in a non-specific cancer population, where treatment decisions are made monthly over a six-month period. At each time $t = 0, \dots, 6$, the patient's state $S_t = (T_t, X_t)$ includes tumor size $T_t$ and treatment toxicity $X_t$. The administered dose $A_t$ ranges from $0$ to $1$ in increments of $0.1$. Tumor and toxicity dynamics are governed by coupled differential equations describing tumor reduction due to treatment, drug toxicity accumulation, and remission once $T_t = 0$. Mortality is modeled by a survival process with hazard $\lambda(t) = \exp(-4 + T_t + X_t)$, reflecting the combined effect of tumor burden and toxicity on survival probability. The reward at each stage $t$ is defined as the sum of three components capturing (i) survival status, (ii) change in toxicity, and (iii) tumor response, with $R_t = R_{t,1} + R_{t,2} + R_{t,3}$, where $R_{t,1} = -60$ if death occurs at stage $t$ and $0$ otherwise, $R_{t,2} = 5$ if $X_{t+1} - X_t \leq -0.5$ and $-5$ otherwise, and $R_{t,3} = 15$ if $T_{t+1} = 0$, $5$ if $T_{t+1} - T_t \leq -0.5$ and $T_{t+1} \neq 0$, and $-5$ otherwise. Each patient trajectory is thus characterized by sequential trade-offs between efficacy and tolerability, providing a suitable environment for evaluating decision rules under multiple near-equivalent treatment sequences.

From this simulation model, we generated a training sample of 500 patients under a randomized treatment strategy and fitted our models using both classical backward Q-learning (Algorithm~\ref{alg:BQ}) and the proposed Near-Equivalent Q-learning procedure (Algorithm~\ref{alg:NearQ}). In both approaches, the same regression model was employed, namely a support vector regression with radial basis function kernel (default $C=1.0$), ensuring comparability between methods.

To illustrate the behavior of the proposed approach under different levels of admissibility, we consider several values of the hyperparameter $\varepsilon \in \{0.1, 0.3, 0.5, 0.9\}$. For each value of $\varepsilon$, the $\varepsilon$-selection criterion is applied within the backward recursion to identify sets of near-equivalent policies.

\begin{figure}[ht]
\centering
\includegraphics[width=0.9\textwidth]{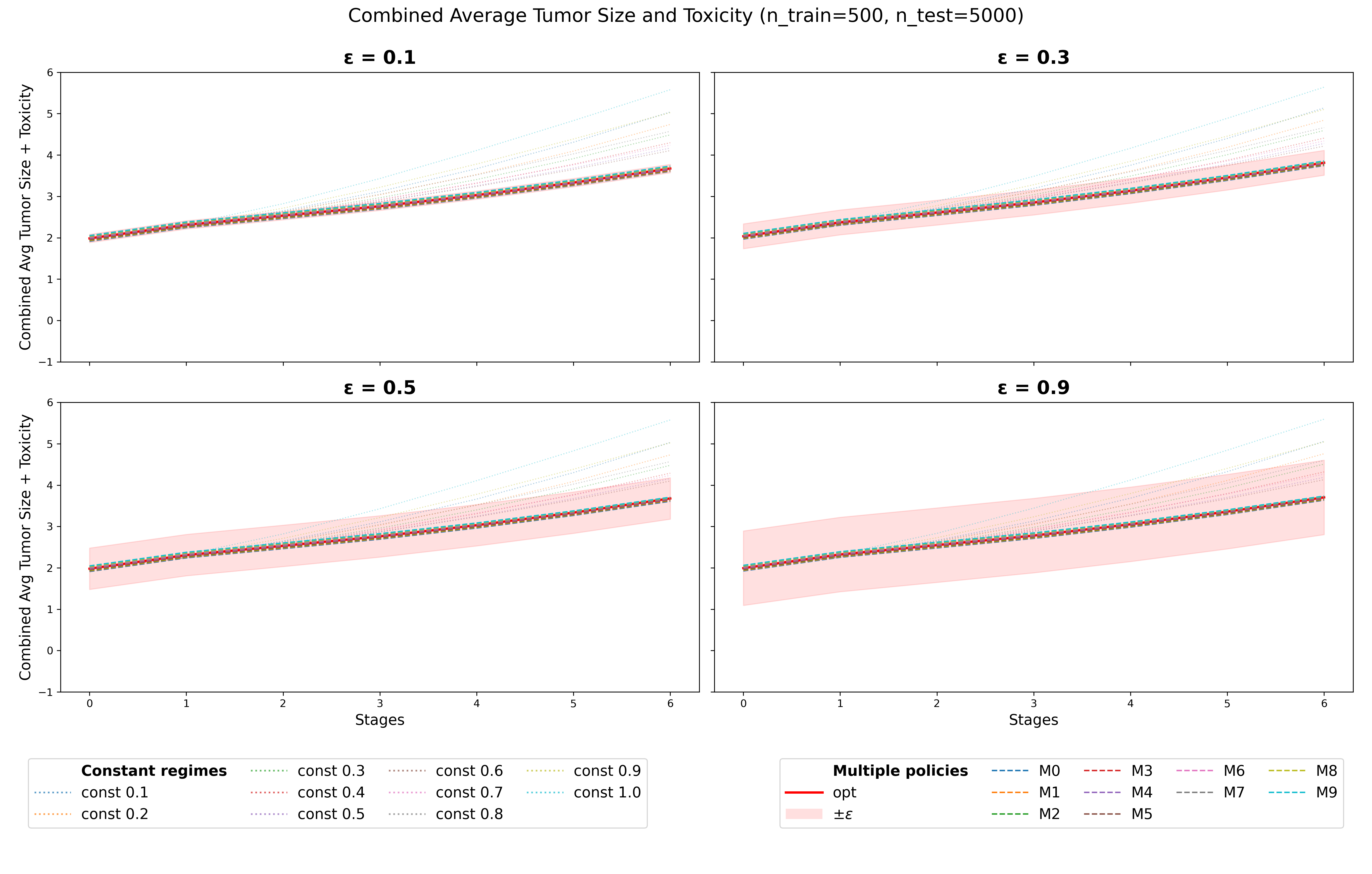}
\caption{Combined average tumor size and toxicity over six stages for 5000 simulated patients under identical initial conditions ($n_{\text{train}}=500$, $n_{\text{test}}=5000$). Each panel corresponds to a different value of $\varepsilon$, illustrating how the admissible set of policies expands as the tolerance increases. Dashed lines correspond to constant treatment regimes ranging from 0.1 to 1.0. The red solid line (``opt'') represents the policy learned via classical backward Q-learning. Bold dashed lines denote the $\varepsilon$-equivalent policies obtained with the proposed method. The shaded red band indicates the $\varepsilon$-tolerance region around the optimal policy.}
\label{fig:cancer}
\end{figure}

The resulting treatment strategies are evaluated by simulating trajectories on an independent test sample of 5000 patients generated under identical initial conditions. Figure~\ref{fig:cancer} displays, over a six-month horizon, the combined average tumor size and toxicity experienced by these patients under the different treatment strategies. The dashed curves correspond to constant treatment regimes ranging from 0.1 to 1.0, while the red solid line (``opt'') represents the policy learned via classical Q-learning. The dashed curves correspond to the near-equivalent policies identified by our method, and the shaded red region illustrates the $\varepsilon$-tolerance band around the optimal trajectory. Each panel corresponds to a different value of the admissibility parameter $\varepsilon$.

All learned policies produce consistently lower values of the combined tumor size–-toxicity metric than the constant treatment regimes across the entire time horizon. Since lower values correspond to a better trade-off between tumor burden and treatment toxicity, this indicates that adaptive treatment strategies achieve more favorable overall outcomes than fixed-dose regimes. The first-ranked near-equivalent policy, in terms of value, perfectly overlaps with the classical optimal policy, confirming that the proposed framework recovers the standard Q-learning solution when $\varepsilon$ is small. As $\varepsilon$ increases, the admissible band gradually widens, reflecting a controlled relaxation of optimality. Nevertheless, the resulting policies remain visually almost indistinguishable from the optimal trajectory and consistently outperform all constant treatment regimes. This indicates that the policy with the highest value within the near-equivalent set remains very similar across values of $\varepsilon$.

This pattern also suggests that the value function exhibits a relatively flat region around the optimum. In other words, several treatment strategies produce nearly identical expected outcomes, indicating that the optimal policy is not isolated but lies within a plateau of near-optimal solutions. The proposed framework explicitly characterizes this region by identifying sets of $\varepsilon$-equivalent policies. Moreover, the performance gap between constant and learned treatment strategies increases over time, reflecting the cumulative impact of sequential decision-making and illustrating the stability of the backward learning process under a controlled relaxation of optimality.

Moreover, the performance gap between constant and learned treatment strategies increases over time, reflecting the cumulative nature of sequential decision-making and illustrating the stability of the backward learning process under a controlled relaxation of optimality. We also report the computational cost of the proposed procedure. All computations were performed in Python on a laptop equipped with an Apple M1 processor (16 GB RAM). Training the classical Q-learning model required approximately $0.29$ seconds, whereas training the near-equivalent Q-learning procedure required about $2.1$ seconds for a given value of $\varepsilon$.

Although the simulation model captures key aspects of dynamic treatment decisions, it remains a simplified representation of the clinical process and is primarily designed to illustrate the methodological behavior of the proposed approach. The limitations of this simulation design and their implications for practical applications are discussed in the next section.

\section{Discussion}


This work introduces a framework for identifying near-equivalent treatment policies within the Q-learning paradigm. While classical Q-learning focuses on estimating a single optimal policy, the proposed approach relaxes this objective by identifying sets of policies whose expected value remains within a controlled tolerance of the optimum. Rather than producing a single deterministic rule, the method characterizes families of treatment strategies that are nearly optimal according to a user-defined admissibility hyperparameter $\varepsilon$. This formulation reflects the practical reality that, in many clinical settings, several treatment options may lead to very similar expected outcomes. Consequently, restricting attention to a single optimal policy may obscure meaningful alternatives and give the impression of a sharply defined decision boundary where little difference in expected benefit actually exists. By explicitly identifying sets of near-optimal strategies, the proposed framework provides a richer representation of treatment decisions and highlights situations in which the optimal choice is inherently unstable.


The hyperparameter $\varepsilon$ controls the tolerance for near-optimality in the treatment recommendation. Conceptually, $\varepsilon$ represents the maximum acceptable loss in expected outcome relative to the optimal treatment. In this work, $\varepsilon$ is interpreted as a tolerance on the difference in expected outcome between competing treatment options, allowing for flexible treatment recommendations in regions where these differences are small.

In applications to clinical data, several practical strategies could guide the choice of $\varepsilon$. When the outcome scale is arbitrary or model-dependent, it may be convenient to define $\varepsilon$ relative to the empirical distribution of the estimated blip function. For instance, $\varepsilon$ could be selected as a fixed fraction of the standard deviation of the blip or as a low quantile of $|\Delta(h)|$. Such scale-relative choices may provide a pragmatic way to calibrate the tolerance level when no natural clinical scale is available. Another possible interpretation is to view $\varepsilon$ as a mechanism for controlling the size of the near-indifference region. By selecting $\varepsilon$ so that a predetermined proportion of individuals satisfies $|\Delta(h)| \le \varepsilon$, one could regulate the fraction of the population for which the treatment recommendation becomes set-valued rather than deterministic. From this perspective, $\varepsilon$ acts as a policy-level tuning parameter governing the trade-off between decisiveness and flexibility. More generally, the choice of $\varepsilon$ would ideally incorporate clinical considerations, since the acceptable deviation from optimality depends on the practical consequences of treatment differences. Future work could therefore explore approaches that combine empirical calibration with clinically meaningful thresholds when selecting $\varepsilon$.

From a statistical standpoint, $\varepsilon$ may also be linked to estimation uncertainty. When uncertainty quantification for the blip function is available, for instance through bootstrap or Bayesian procedures, $\varepsilon$ may be chosen to reflect the typical magnitude of estimation error. Treatments are then considered near-optimal when differences in estimated $Q$-values cannot be distinguished given the uncertainty of the model. Interpreting $\varepsilon$ as a tolerance on instantaneous regret therefore provides a unified perspective linking clinical relevance, statistical uncertainty, and reinforcement learning principles.


Several design choices were considered when integrating the $\varepsilon$-selection step within the backward Q-learning recursion. In the proposed implementation, the selection is performed at the penultimate stage of the algorithm. Introducing the criterion at the final stage would effectively impose the admissibility condition on a transformed version of the final outcome, which could lead to cumulative drift when propagated backward through the regression steps.

Conversely, introducing the $\varepsilon$-selection too early in the recursion would collapse the decision process to a single policy in subsequent stages due to the structure of the backward regression updates. The penultimate stage therefore represents a natural compromise, allowing the admissible set of policies to expand while preserving a direct connection with the final objective.

We also considered applying the $\varepsilon$-selection criterion at every stage of the backward recursion. Such an approach would generate a branching structure of candidate policies whose size grows exponentially with the number of stages and the cardinality of the action space. In addition to the resulting computational burden, many of the resulting policies would be identical, making the criterion insufficiently restrictive in practice.


From the perspective of individualized treatment rules illustrated in Section~4, the admissibility criterion naturally defines regions of treatment indifference. When these regions are large, the resulting policy may recommend multiple treatments for a given patient because the estimated treatment effects are very similar. In such situations, enforcing a single treatment recommendation may be arbitrary and overly sensitive to estimation noise. The proposed framework instead explicitly acknowledges this ambiguity by identifying sets of $\varepsilon$-optimal treatments. Consequently, several treatments may reasonably be considered acceptable, allowing external considerations such as patient preference, treatment accessibility, cost, or safety to inform the final decision. More broadly, the width of the admissible region can be interpreted as an indicator of treatment indifference: narrow regions correspond to clear treatment preference, whereas wider regions indicate that multiple treatments yield nearly identical expected outcomes. In this way, the proposed method provides additional insight beyond a single optimal rule by characterizing the stability of the treatment decision.


When considered in the context of dynamic treatment regimes presented in Section~5, the proposed framework identifies sets of sequential policies whose cumulative value remains within an $\varepsilon$-neighborhood of the optimal policy. Rather than focusing exclusively on a single optimal sequence of decisions, the method characterizes families of near-optimal strategies that produce similar long-term outcomes. This perspective is particularly relevant in sequential decision problems where treatment effects accumulate over time, as small deviations from the optimal decision at a given stage may have only limited impact on the overall outcome, leading to multiple policies with very similar value. Identifying such policies therefore provides insight into the stability of the sequential decision process and may facilitate the integration of practical constraints or expert judgment into the final treatment strategy. The simulation setting considered in this work is intentionally simplified and does not fully reflect the complexity typically encountered in precision medicine applications. In particular, the state space includes only two covariates describing tumor size and toxicity, which limits the heterogeneity of patient profiles and reduces the potential for highly individualized treatment recommendations. Moreover, patient trajectories are generated through a deterministic dynamical system based on ordinary differential equations. Although this formulation captures interactions between tumor progression and treatment toxicity, it also produces relatively smooth and structured trajectories, resulting in limited variability in the state space and potentially contributing to the relatively stable set of near-equivalent policies observed in the simulation results. These simplifications were deliberately adopted to provide a controlled environment in which the methodological behavior of the proposed framework can be clearly illustrated. In more realistic clinical settings involving higher-dimensional patient histories and more complex stochastic dynamics, the structure of near-equivalent policies may be substantially richer.


The additional computational cost induced by the proposed framework remains moderate and primarily reflects the identification of multiple admissible treatment strategies during the backward recursion. In practice, the number of near-equivalent strategies that may be explored is naturally bounded by the size of the action space, so the complexity of the procedure remains directly linked to the number of available treatment options.


Several extensions of the proposed framework deserve further investigation. First, uncertainty quantification for near-equivalent treatment strategies could be developed using bootstrap procedures applied to the entire learning process. Such an approach would allow the construction of confidence measures for the admissible treatment sets and provide additional insight into the stability of the recommended strategies. However, this task is complicated by the presence of near-equivalent treatment options, which naturally raises questions of non-identifiability, as small differences in estimated value may not be statistically distinguishable. In such regions, the decision rule may exhibit non-smooth behavior, leading to non-standard asymptotics and potentially limiting the validity of standard bootstrap procedures. Addressing these challenges may require the development of alternative inferential tools tailored to set-valued or near-indifferent decision problems.

Second, the current implementation estimates a separate regression model for each action. An interesting extension would consist in estimating the Q-function through vector-valued or multi-output regression models, allowing information to be shared across treatment options. Related ideas have been explored in the context of joint or shared-parameter estimation approaches \citep{wang2022adaptive}, which aim to improve statistical efficiency by leveraging common structure across treatments. Such approaches could potentially enhance estimation when value functions associated with different treatments are similar. In this setting, because only the outcome corresponding to the treatment actually received is observed for each individual, multi-output formulations would require careful modeling to account for this partially observed structure and to avoid introducing bias in the estimation of treatment effects.

Finally, the proposed $\varepsilon$-selection framework is not restricted to classical Q-learning and could potentially be combined with other regression-based methods for estimating individualized treatment rules, such as dynamic weighted ordinary least squares or G-estimation. Applying the approach to real-world datasets involving multiple treatment alternatives represents an important direction for future work.



\section*{Acknowledgments}
Erica E. M. Moodie is a Canada Research Chair (Tier 1) in Statistical Methods for Precision Medicine, funded by the Canadian Institutes of Health Research (CIHR). This work was supported by a Discovery Grant from the Natural Sciences and Engineering Research Council of Canada (NSERC). Computational resources were provided by the Digital Research Alliance of Canada (Alliance Canada).

\section*{Financial disclosure}

None reported.

\section*{Conflict of interest}

The authors declare no potential conflict of interests.

\bibliographystyle{apalike}
\bibliography{references}

@article{zhao2009reinforcement,
  title = {Reinforcement learning design for cancer clinical trials},
  author = {Zhao, Yufan and Kosorok, Michael R. and Zeng, Donglin},
  journal = {Statistics in Medicine},
  volume = {28},
  number = {26},
  pages = {3294--3315},
  year = {2009},
  doi = {10.1002/sim.3707}
}

@article{furnkranz2012preference,
  title = {Preference-based reinforcement learning: A formal framework and a policy iteration algorithm},
  author = {F{\"u}rnkranz, Johannes and H{\"u}llermeier, Eyke and Cheng, Weiwei and Park, Sang-Hyeun},
  journal = {Machine Learning},
  volume = {89},
  pages = {123--156},
  year = {2012},
  doi = {10.1007/s10994-012-5317-9}
}

@inproceedings{akrour2012april,
  title = {APRIL: Active preference learning-based reinforcement learning},
  author = {Akrour, Riad and Schoenauer, Marc and Sebag, Mich{\`e}le},
  booktitle = {Proceedings of the European Conference on Machine Learning and Knowledge Discovery in Databases (ECML PKDD)},
  pages = {116--131},
  year = {2012},
  doi = {10.1007/978-3-642-33460-3_10}
}

@phdthesis{humphrey2017using,
  title = {Using reinforcement learning to personalize dosing strategies in a simulated cancer trial with high dimensional data},
  author = {Humphrey, Kyle},
  school = {The University of Arizona},
  year = {2017}
}

@article{goldberg2012q,
  title = {Q-learning with censored data},
  author = {Goldberg, Yair and Kosorok, Michael R.},
  journal = {Annals of Statistics},
  volume = {40},
  number = {1},
  pages = {529--560},
  year = {2012},
  doi = {10.1214/12-AOS971}
}

@article{clifton2020q,
  title = {Q-learning: Theory and applications},
  author = {Clifton, Jesse and Laber, Eric},
  journal = {Annual Review of Statistics and Its Application},
  volume = {7},
  number = {1},
  pages = {279--301},
  year = {2020},
  doi = {10.1146/annurev-statistics-031219-041209}
}

@article{fard2011non,
  title = {Non-deterministic policies in Markovian decision processes},
  author = {Fard, Mohammad Milani and Pineau, Joelle},
  journal = {Journal of Artificial Intelligence Research},
  volume = {40},
  pages = {1--24},
  year = {2011},
  doi = {10.1613/jair.3110}
}

@inproceedings{tang2020clinician,
  title = {Clinician-in-the-loop decision making: Reinforcement learning with near-optimal set-valued policies},
  author = {Tang, Shengpu and Modi, Aditya and Sjoding, Michael and Wiens, Jenna},
  booktitle = {Proceedings of the International Conference on Machine Learning (ICML)},
  pages = {9387--9396},
  year = {2020},
  organization = {PMLR}
}

@article{lizotte2016multi,
  title = {Multi-objective Markov decision processes for data-driven decision support},
  author = {Lizotte, Daniel J. and Laber, Eric B.},
  journal = {Journal of Machine Learning Research},
  volume = {17},
  number = {210},
  pages = {1--28},
  year = {2016}
}

@book{chakraborty2013statistical,
  title = {Statistical methods for dynamic treatment regimes},
  author = {Chakraborty, Bibhas and Moodie, Erica E. M.},
  volume = {2},
  year = {2013},
  publisher = {Springer},
  doi = {10.1007/978-1-4614-7428-9}
}

@book{kosorok2015adaptive,
  title = {Adaptive treatment strategies in practice: Planning trials and analyzing data for personalized medicine},
  author = {Kosorok, Michael R. and Moodie, Erica E. M.},
  year = {2015},
  publisher = {SIAM},
  doi = {10.1137/1.9781611973860}
}

@article{yazzourh2025medical,
  title = {Medical knowledge integration into reinforcement learning algorithms for dynamic treatment regimes},
  author = {Yazzourh, Sophia and Savy, Nicolas and Saint-Pierre, Philippe and Kosorok, Michael R.},
  journal = {International Statistical Review},
  year = {2025},
  doi = {10.1111/insr.12617}
}

@article{kosorok2019precision,
  title={Precision medicine},
  author={Kosorok, Michael R and Laber, Eric B},
  journal={Annual review of statistics and its application},
  volume={6},
  number={1},
  pages={263--286},
  year={2019},
  publisher={Annual Reviews}
}

@incollection{robins2000marginal,
  title={Marginal Structural Models Versus Structural Nested Models as Tools for Causal Inference},
  author={Robins, James M},
  booktitle={Statistical Models in Epidemiology, the Environment, and Clinical Trials},
  pages={95--133},
  year={2000},
  publisher={Springer}
}

@article{robins1989analysis,
  title={The Analysis of Randomized and Non-Randomized AIDS Treatment Trials Using a New Approach to Causal Inference in Longitudinal Studies},
  author={Robins, James M},
  journal={Health Service Research Methodology: a Focus on AIDS},
  pages={113--159},
  year={1989},
  publisher={US Public Health Service}
}

@article{robins1992estimation,
  title={Estimation of the Time-Dependent Accelerated Failure Time Model in the Presence of Confounding Factors},
  author={Robins, James M},
  journal={Biometrika},
  volume={79},
  number={2},
  pages={321--334},
  year={1992},
  publisher={Oxford University Press}
}

@article{robins1998correction,
  title={Correction for Non-Compliance in Equivalence Trials},
  author={Robins, James M},
  journal={Statistics in Medicine},
  volume={17},
  number={3},
  pages={269--302},
  year={1998},
  publisher={Wiley Online Library}
}

@ARTICLE{Orellana2010-zu,
  title    = "Dynamic Regime Marginal Structural Mean Models for Estimation of Optimal Dynamic Treatment Regimes, Part I: Main Content",
  author   = "Orellana, Liliana and Rotnitzky, Andrea and Robins, James M",
  journal  = "The International Journal of Biostatistics",
  volume   =  6,
  number   =  2,
  pages    = "Article 8",
  year     =  2010
}

@ARTICLE{Zhao2012-vc,
  title    = "Estimating Individualized Treatment Rules Using Outcome Weighted
              Learning",
  author   = "Zhao, Ying-Qi and Zeng, Donglin and Rush, A John and Kosorok,
              Michael R",
  journal  = "Journal of the American Statistical Association",
  volume   =  107,
  number   =  449,
  pages    = "1106--1118",
  month    =  sep,
  year     =  2012,
}

@article{wallace2015doubly,
  title={Doubly-robust dynamic treatment regimen estimation via weighted least squares},
  author={Wallace, Michael P and Moodie, Erica EM},
  journal={Biometrics},
  volume={71},
  number={3},
  pages={636--644},
  year={2015},
  publisher={Oxford University Press}
}

@article{song2015penalized,
  title={Penalized q-learning for dynamic treatment regimens},
  author={Song, Rui and Wang, Weiwei and Zeng, Donglin and Kosorok, Michael R},
  journal={Statistica Sinica},
  volume={25},
  number={3},
  pages={901},
  year={2015}
}

@article{zhao2015new,
  title={New statistical learning methods for estimating optimal dynamic treatment regimes},
  author={Zhao, Ying-Qi and Zeng, Donglin and Laber, Eric B and Kosorok, Michael R},
  journal={Journal of the American Statistical Association},
  volume={110},
  number={510},
  pages={583--598},
  year={2015},
  publisher={Taylor \& Francis}
}

@article{zhou2017residual,
  title={Residual weighted learning for estimating individualized treatment rules},
  author={Zhou, Xin and Mayer-Hamblett, Nicole and Khan, Umer and Kosorok, Michael R},
  journal={Journal of the American Statistical Association},
  volume={112},
  number={517},
  pages={169--187},
  year={2017},
  publisher={Taylor \& Francis}
}

@PHDTHESIS{Watkins1989-er,
  title    = "Learning from Delayed Rewards",
  author   = "Watkins, Christopher John Cornish",
  editor   = "Young, Richard",
  year     =  1989,
  school   = "King's College",
}

@article{watkins1992q,
  title={Q-Learning},
  author={Watkins, Christopher JCH and Dayan, Peter},
  journal={Machine Learning},
  volume={8},
  pages={279--292},
  year={1992},
  publisher={Springer}
}

@phdthesis{yazzourh2024reinforcement,
  title={Reinforcement learning and Bayesian outcome-weighted learning for precision medicine: integrating medical knowledge into decision-making algorithms},
  author={Yazzourh, Sophia},
  year={2024},
  school={Universit{\'e} de Toulouse}
}

@article{holzinger2016interactive,
  title={Interactive Machine Learning for Health Informatics: When Do We Need the Human-in-the-Loop?},
  author={Holzinger, Andreas},
  journal={Brain Informatics},
  volume={3},
  number={2},
  pages={119--131},
  year={2016},
  publisher={Springer}
}

@article{holzinger2019causability,
  title={Causability and Explainability of Artificial Intelligence in Medicine},
  author={Holzinger, Andreas and Langs, Georg and Denk, Helmut and Zatloukal, Kurt and M{\"u}ller, Heimo},
  journal={Wiley Interdisciplinary Reviews: Data Mining and Knowledge Discovery},
  volume={9},
  number={4},
  pages={e1312},
  year={2019},
  publisher={Wiley Online Library}
}

@article{maadi2021review,
  title={A Review on Human-AI Interaction in Machine Learning and Insights for Medical Applicationss},
  author={Maadi, Mansoureh and Akbarzadeh Khorshidi, Hadi and Aickelin, Uwe},
  journal={International Journal of Environmental Research and Public Health},
  volume={18},
  number={4},
  pages={2121},
  year={2021},
  publisher={MDPI}
}

@article{love2023should,
  title={Who Should I Trust? Cautiously Learning with Unreliable Experts},
  author={Love, Tamlin and Ajoodha, Ritesh and Rosman, Benjamin},
  journal={Neural Computing and Applications},
  volume={35},
  number={23},
  pages={16865--16875},
  year={2023},
  publisher={Springer}
}

@article{wang2022adaptive,
  title={Adaptive treatment strategies for chronic conditions: shared-parameter G-estimation with an application to rheumatoid arthritis},
  author={Wang, Shouao and Moodie, Erica Em and Stephens, David A and Nijjar, Jagtar S},
  journal={Biostatistics},
  volume={23},
  number={2},
  pages={430--448},
  year={2022},
  publisher={Oxford University Press}
}

@article{chakraborty2010inference,
  title={Inference for non-regular parameters in optimal dynamic treatment regimes},
  author={Chakraborty, Bibhas and Murphy, Susan and Strecher, Victor},
  journal={Statistical methods in medical research},
  volume={19},
  number={3},
  pages={317--343},
  year={2010},
  publisher={SAGE Publications Sage UK: London, England}
}

@article{moodie2010estimating,
  title={Estimating optimal dynamic regimes: Correcting bias under the null},
  author={Moodie, Erica EM and Richardson, Thomas S},
  journal={Scandinavian Journal of Statistics},
  volume={37},
  number={1},
  pages={126--146},
  year={2010},
  publisher={Wiley Online Library}
}

\end{document}